# Empowering OLAC Extension using *Anusaaraka* and Effective text processing using Double Byte coding

*Dr. B Prabhulla Chandran Pillai,*
*Catholicate College,*
*(Affiliated to Mahatma Gandhi University, Kottayam),*
*Pathanamthitta, Kerala, India[1]*

## Abstract

The paper reviews the hurdles while trying to implement the OLAC extension for Dravidian / Indian languages. The paper further explores the possibilities which could minimise or solve these problems. In this context, the Chinese system of text processing and the *anusaaraka* system are scrutinised.

**Keywords:** Indian language, OLAC, Text Processing, Machine Translation, *Anusaaraka*

## 1 Introduction

Though the OLAC extension for Dravidian languages (can be extended to all Indian languages) was proposed (*B Prabhulla Chandran Pillai*, 2009), it is may not be effective if we don't address the issues with regard to the Machine Translation (MT) process and the processing of Indian languages. For example the current architecture won't support *the multiplatform multilingual system*, in the context of Indian languages. This can be comprehended well if we look at it from a *Computational Linguistic* point of view.

*Anusaaraka* system implementations simplify the problems associated with the MT process. Also by supporting a commonsense based approach, the system can be improvised further ((*B Prabhulla Chandran Pillai*, 2009). The issues associated with the inefficiency in text processing can be solved to a considerable extend by following the Chinese text processing systems (*Maruf Hasan, 1995*) which use *Double Byte Coding*. Since Indian Languages (especially the Sanskrit based ones) have been using *Single Byte* coding for text processing, the processing capabilities (of current method) are low.

## 2. Review

The non-availability of lexical resources is a potential challenge for any researcher whose works are pivoted around the field of NLP on Indian languages. Many proposals like the developing of a bilingual electronic dictionary (mostly done as a collaborative effort) helped the researches to fix the issue to a good extend. Later, with the proposal (and implementation) of *anusaaraka system* this problem was addressed properly.

Building a fully-automatic general purpose high quality machine translation system (FGHMT) is not a practically viable solution. So the only option is to improvise the proposal of *anusaaraka-like* systems.

Further, Sanskrit based Indian Languages are relatively difficult for a computer to process owing to their two dimensional features. But a simple analysis of the *Chinese Information Processing Technique* shows that their method is quite efficient. This force us to model a system for Sanskrit based Indian Languages, which is akin to the *Chinese Information Processing Technique*. This can be done on a *Multilingual System Architecture* (*Maruf Hasan, 1995*) as illustrated below.

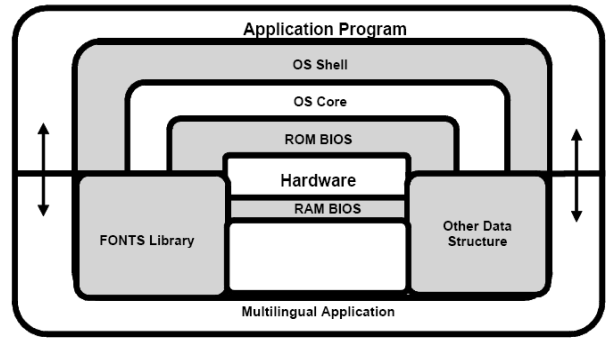

---

[1] Email: bprabhullachandran@gmail.com

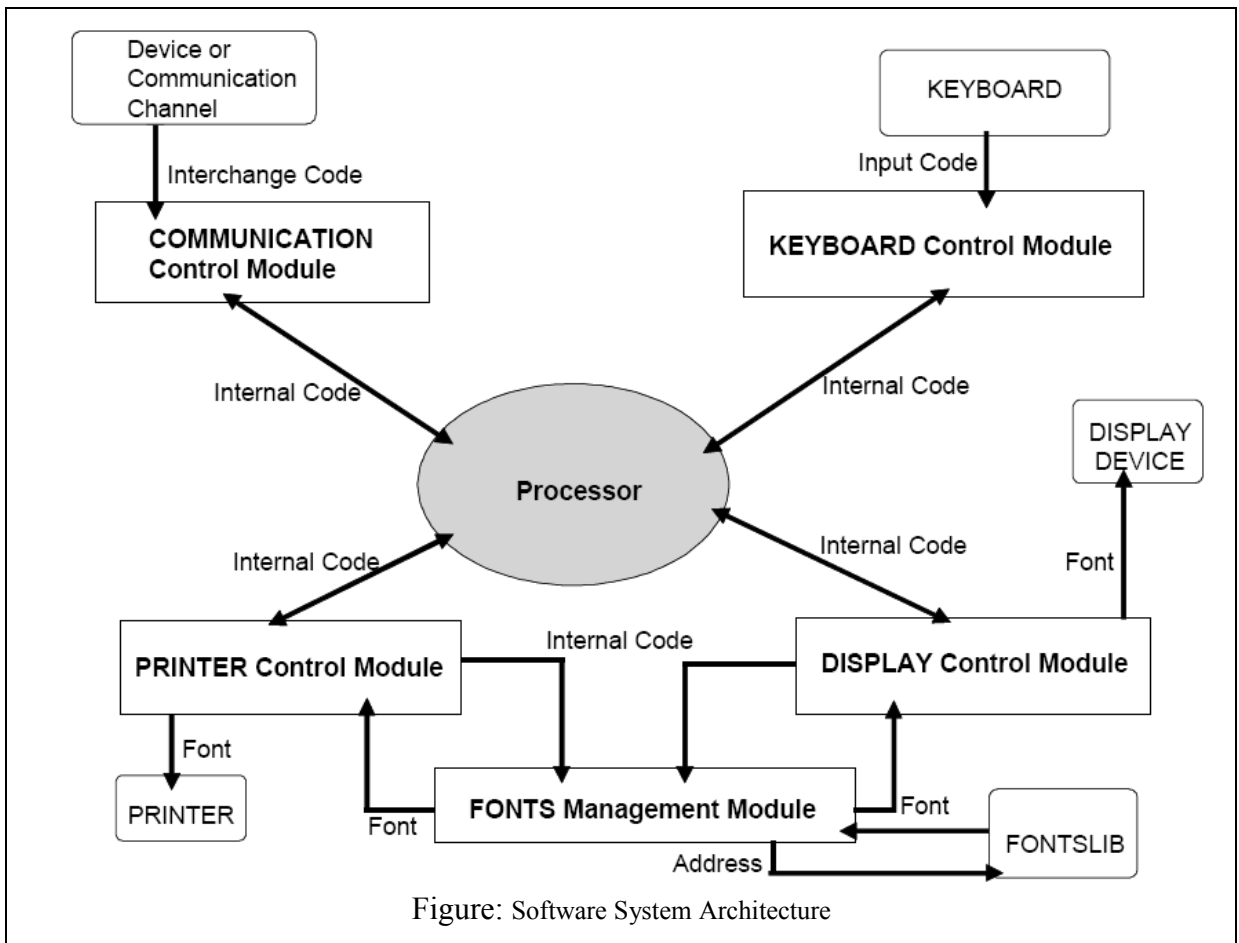

Figure: Software System Architecture

Indian languages are not processed effectively by computers, except for word processing purpose. This situation will change if we follow *Double Byte Coding* just like the case where a *double byte code* (called as the *Internal Code*) is used for every Chinese ideographic character. '*Software System Architecture figure*' allows us to comprehend the architectural changes demanded by the system.

3. Analysis and Suggestions

A 16x16 bitmapped font (primarily for display) is made and saved as a binary file for representing a given Chinese ideographic character. And for printing 24x24, 48x48 bitmapped fonts are used. And for receiving the input various schemas are used, out of which the *PinYin-HanZi* conversion input method is very popular. This method gives a keyboard layout to the user using the phonetic alphabet. A table of input keys and Chinese character code is used for mapping the corresponding Chinese character.

*Display Control Module* gives the address of the corresponding ideographic character in the fonts library and retrieve the same from it using *Font Management Module*. Later this is passed onto the display device for displaying. Communication protocols are also supported by converting the *Internal Code* into the *Interchange Code*.

Since most of the Indian languages are the derivative of *Sanskrit,* they have a lot in common. *Hindi, Bengali Nepali, Tamil, Gujarati, Oriya, Telegu, Kannada, Malayalam* etc are among them. Among these, *Hindi and Nepali* are written in Devanagari script itself. But by looking at the '*comparison figure*' (between Devanagari and Bengali), one can understand how closely the languages are related (even though the scripts are different). This is true for all the above mentioned languages.

The complexity of the languages arises due to multiple reasons. For example; consonants and vowels, consonants and consonants may combine

*Figure: Comparison of Devanagari and Bengali*

together and that form may look differently. In some cases, vowels wrap its host consonant. Glyphs of the letters (and their variations) are employed by most of the systems for making their font set as shown below:

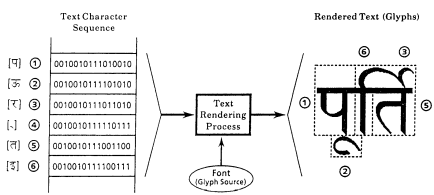

Thus the user will have to use a keyboard layout which is based on glyphs, to provide the input.

This can be substituted with a *Double Byte Coding* ((*Maruf Hasan, 1995):*

*"The system is more like a CCDOS4 (Chinese Character Disk Operating System) system. CCDOS encoded a total of 6763 Chinese characters. According to GB2312-80 , the first level Chinese character consist of 3755 characters which cover 95% character of modern Chinese text. The next level consists of another 3008 Chinese characters*

*which cover another 5% of modern Chinese text. As an experiment, we just replaced the Second Level character fonts from CCLIB (CCDOS fonts library) with Bengali character fonts. Then we designed an External Input Module5 where we used a new conversion table and search algorithm to manipulate bilingual input. Output (especially display and print) mechanism needs no modification. This system successfully provided multilingual (Bengali, Chinese and English) environment in a PC. Other Localized Chinese software like Chinese Dbase, Chinese Word Star also supported the multilingual environment."*

By following such type of an approach we can have a *Global Text Processing System* for Indian Languages.

Even if we succeed in effectively processing this, the *OLAC* extension is not completely feasible unless we figure out a way by which the text in one Indian language accessible in another Indian language. As we said, this can be done using *anusaaraka* system. This system also employs the methodology wherein we incorporate the common sense knowledge in processing the language. Here the load is shared between man and computer; in such a way that the language load is taken care of by the machine and the interpretation of the text is done by man.

The free-order of words is a feature of any Indian language and '*information that relates an action (verb) to its participants (nouns) is primarily expressed by means of post-positions or case endings of nouns (collectively called vibhaktis of the noun)'.* (ANUSAARAKA: OVERCOMING THE LANGUAGE BARRIER IN INDIA, 2001)

The paper also gives an example that illustrates the mode of working of the *anusaaraka* system

```
rAma ne roTI khAI
Ram-erg. bread ate
Ram ate the bread.
```

*The ergative (erg.) post postion marker ('ne') after 'rAma' indicates that Ram is the \*karta\* of eat, which here means that Ram is the \*agent\* of eating*

If the 'reader' (*processing part*) is *human*, then he knows that *Ram* is the name of the individual and '*bread'* is for eating. But as far as the computer is concerned both these nouns are same. Still we can process the language to a great extend by morphological processing and using bilingual dictionary. But a human being may also perform a language related analysis or generation, which will make the translation untruthful to the original text. This can be solved by employing the *anusaaraka* system, as shown above.

Here is another example that shows a similar *action*:

T: mIru pustakaM caduvutunnArA?
@H: Apa pustaka paDha_raHA_[HE| thA]_kyA{23_ba.}?
!E: You book read_ing_[is|was]_Q.?
E: Are/were you reading a book?

(Where T=Telugu, @H=anusaaraka Hindi, !E=English gloss, E=English)

In this way, inter-conversion between Indian languages can be done effectively.

There are even many algorithms in Artificial Neural Network (ANN) that can help the machine to learn ideas. We can also employ Agent based systems (ABS) for assisting in machine learning.

**4.** Conclusion and Remarks

Though processing of the languages and MT are two difficult tasks that makes OLAC extension difficult, those issues can be solved by adopting a *Double Byte Coding* for the language and using *anusaaraka system.* Further this can be improvised by employing the ANN and ABS algorithms.

Thus a researcher can easily extend OLAC so as to enable processing and preservation (using MT) of Indian languages; rather than just being an index for finding data, tools and resources. And OLAC can use its present index structure to do (and organize) this *must-do* task effectively.

Also, in order to get best results, we can categorize Indian languages in the following way:

1. Northern languages (Punjabi, Kashmiri, Urdu, etc)
2. Western languages (Konkani, Marathi, Gujarati, etc.)
3. South Indian languages (Tamil, Telugu, Kannada, Malayalam, etc.)
4. Eastern languages (Bengali, Assamese, Oriya, etc.)

5. References

*B Prabhulla Chandran Pillai*. 2009. An OLAC Extension for Dravidian Languages. arXiv:cs/0908.4431v1

The People's Republic of China National Standard code of Chinese Graphic Character Set for Information Interchange, Primary Set, GB2312-80, Beijing, China, 1981.

Chinese Code for Data Communication, Ministry of Communications, Republic of China, August 1983.

The Unicode Consortium, The Unicode Standard, Version 1.0, Volume 1 & 2, Addison Wesley, 1990-2.

Bharati, Akshar, Vineet Chaitanya, Rajeev Sangal, Natural Language Processing: A Paninian Perspective, Prentice-Hall of India, 1995.

Sharma, Dipti M, Building Lexical Resources, in Proc. of Symposium in Information Revolution in Indian Languages,Osmania University, 13-15 November 1999.

Bharati, Akshar, Dipti M Sharma, Rajeev Sangal, TransLexGram : An Introduction,Technical Report no: TR-LTRC-011, LTRC, IIIT Hyderabad, Jan 2001.

Bharati, Akshar, Dipti M Sharma, Rajeev Sangal, TransLexGram : Guidelines for Verb Frames, Technical Report no: TR-LTRC-013, LTRC, IIIT Hyderabad, Jan 2001.

Bharati, Akshar, Dipti M Sharma, Rajeev Sangal, AnnCorra : An Introduction, Technical Report no: TR-LTRC-014, LTRC, IIIT Hyderabad, Mar 2001.

LRNLP-2001: Workshop on Lexical Resources for Natural Language Processing for Indian Languages, Hyderabad, January 2001.

Shabdaanjali: English - Hindi e-Dictionary ver.0.2, 2000

Natural Language Processing: A Paninian Perspective, Akshar Bharati, Vineet Chaitanya, Rajeev Sangal, Prentice-Hall of India, 1995

Anusaraka: A Device to Overcome the Language Barrier, V.N. Narayana, Ph.D. thesis, Dept. of CSE, I.I.T. Kanpur, 1994.